\documentclass[letterpaper, 10 pt, conference]{ieeeconf}  

\IEEEoverridecommandlockouts                              

\usepackage{cite}
\usepackage{amsmath,amssymb,amsfonts}
\usepackage{algorithmic}
\usepackage{graphicx}
\usepackage{textcomp}
\usepackage{xcolor}
\usepackage{booktabs}
\usepackage{hyperref}
\usepackage{multirow}

\usepackage{cleveref} 

\title{\LARGE \bf
Task-Driven Manipulation with Reconfigurable Parallel Robots
}

\author{Daniel Morton$^{1}$, Mark Cutkosky$^{1}$, and Marco Pavone$^{2}$
\thanks{Support for this work was provided by NASA under the NIAC program (Grant \#80NSSC22K0766). D. Morton is supported by NSF GRFP.}
\thanks{$^{1}$Daniel Morton and Mark Cutkosky are with the Department of Mechanical Engineering, Stanford University, Stanford, CA 94305.
        {\tt\small \{dmorton, cutkosky\}@stanford.edu}}%
\thanks{$^{2}$Marco Pavone is with the Department of Aeronautics and Astronautics, Stanford University, Stanford, CA 94305.
        {\tt\small pavone@stanford.edu}}%
}

\begin{document}

\maketitle
\thispagestyle{empty}
\pagestyle{empty}

\begin{abstract}

ReachBot, a proposed robotic platform, employs extendable booms as limbs for mobility in challenging environments, such as martian caves. When attached to the environment, ReachBot acts as a parallel robot, with reconfiguration driven by the ability to detach and re-place the booms. This ability enables manipulation-focused scientific objectives: for instance, through operating tools, or handling and transporting samples. To achieve these capabilities, we develop a two-part solution, optimizing for robustness against task uncertainty and stochastic failure modes. First, we present a mixed-integer stance planner to determine the positioning of ReachBot's booms to maximize the task wrench space about the nominal point(s). Second, we present a convex tension planner to determine boom tensions for the desired task wrenches, accounting for the probabilistic nature of microspine grasping. We demonstrate improvements in key robustness metrics from the field of dexterous manipulation, and show a large increase in the volume of the manipulation workspace. Finally, we employ Monte-Carlo simulation to validate the robustness of these methods, demonstrating good performance across a range of randomized tasks and environments, and generalization to cable-driven morphologies. We make our code available at our project webpage, {\small\url{https://stanfordasl.github.io/reachbot_manipulation/}}

\end{abstract}


\section{Introduction}

Scientific research missions in space have significantly advanced our understanding of Earth, the solar system, and the broader cosmos. Recently, caves and lava tubes on celestial bodies like the Moon and Mars have emerged as areas of interest due to their distinct geological and astrobiological characteristics, offering potential insights into the history of the solar system \cite{cave_science,nasa_science}. Further exploration of these regions may uncover details about their past habitability, but this exploration is limited by the capabilities of existing robotic platforms \cite{perseverance}, and their inability to access hard-to-reach locations. While passive observation of the environment is suitable for some missions, direct interaction and manipulation unlock a broader range of scientific possibilities, particularly when unconstrained by a small workspace. For instance, a robot with broad mobility and manipulation capabilities could extract geological samples from challenging regions, or assemble preliminary infrastructure for a habitat, in preparation for human arrival.

\begin{figure}
    \centering
    \includegraphics[width=\linewidth]{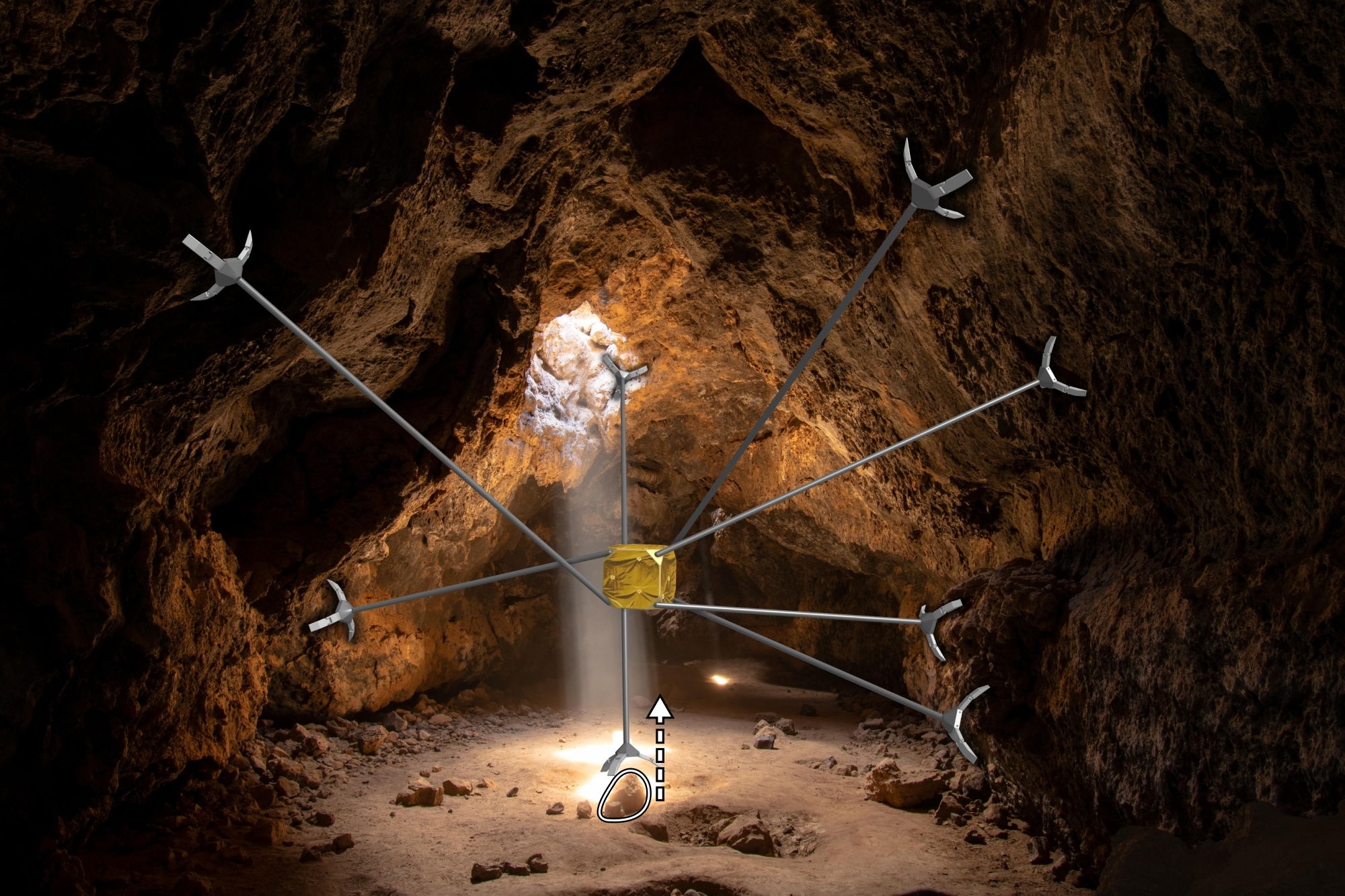}
    \caption{ReachBot performs a sample-extraction manipulation task in a cave environment. By optimally planning (i) where to place the booms and (ii) the tensions in each boom, we ensure that this task is completed even with uncertain pose and wrench estimation, and stochastic failure modes.}
    \label{fig:rb_manip}
    \vspace{-4mm}
\end{figure}

To address this need, ReachBot is a proposed robotic concept for enhanced mobility in challenging environments \cite{reachbot_main,reachbot_design,reachbot_planning,reachbot_hardware,reachbot_sciro}. Using deployable booms as reconfigurable prismatic joints, ReachBot can extend and traverse across large regions, accessing hard-to-reach areas of scientific interest (Fig. \ref{fig:rb_manip}). However, accessing these areas is only the first step of a larger science mission with manipulation-based objectives. Building on the sample extraction example, this task will require (i) aligning a camera to analyze a region while maintaining a steady view, (ii) applying a force or torque to extract the sample, through grasping or drilling, and (iii) transporting the sample to a retrieval location. During this process, ReachBot must also address the challenges of (i) the stochastic nature of grasping a rock with microspines, (ii) uncertainty in the required set of wrenches and poses, and (iii) vibrations and disturbances introduced during execution.

\subsection{Related Work}

\begin{figure*}[t]
    \centering
    \smallskip
    \includegraphics[width=\textwidth]{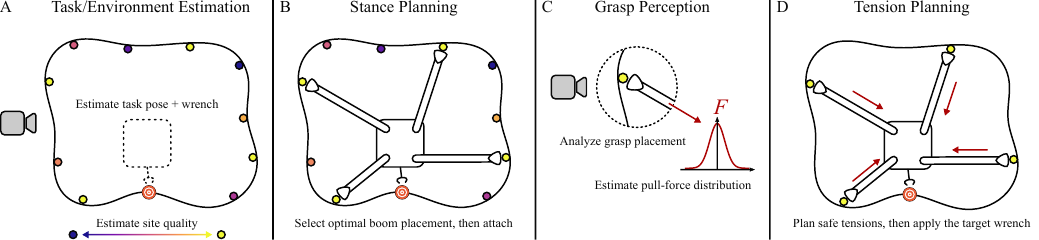}
    \caption{\textbf{Concept of operations.} (A): Upon approaching an area of scientific interest, ReachBot's perception system identifies where the task needs to be performed, estimates a task wrench, identifies candidate grasp sites in the area, and estimates the quality of each site. (B): The \textit{stance planner} considers the grasp sites and determines the optimal placement of each boom, for robust task execution. (C): After each boom is attached, the perception system can identify the quality of each grasp and determine the expected pull-force distribution for each site. (D): Using this grasp quality information, the \textit{tension planner} determines the tensions on each boom to safely achieve the target task wrench.}
    \label{fig:ops}
\end{figure*}


For this manipulation application, with multiple booms attached to the environment, ReachBot is most directly related to cable-driven parallel robots (CDPRs) and cable-suspended robots (CSRs) \cite{cdpr_review}. Of these, ReachBot is comparable to fully-constrained 6D (position and orientation) CDPRs (requiring at least 7 cables) -- as opposed to CDPRs which only control position with fewer than 7 cables. Representative examples of fully-constrained and under-constrained CDPRs are \cite{ipanema} and \cite{skycam}, respectively.

Generally, these CDPRs are not reconfigurable, meaning that the positions of the cable attachment points are fixed. There are a few notable examples of reconfigurable CDPRs \cite{reconfigurable_cdprs}, yet often this reconfigurability is either limited to a single translational axis, or it is a process that is not performed on a task-by-task basis. SpiderBot \cite{spiderbot} is an example of a highly reconfigurable CSR, and is likely the best direct comparison to ReachBot. However, (i) SpiderBot is not fully-constrained, having only 4 cables, and (ii) the authors only study mobility planning rather than using it for manipulation. Others have considered a task-specific optimization for CDPRs \cite{cdpr_task_opt} with similar objectives such as maximizing the tension-closure workspace volume, or minimizing the cable force. However, this analysis is limited to a 3-DOF universal joint and does not consider reconfigurability.

Due to ReachBot's similarities to dexterous manipulation (trading fingers that push for booms that pull) \cite{reachbot_sciro}, we can also look at grasp optimization, much of which builds on the work of \cite{ferrari_canny}. Of particular interest are methods for planning over a set of discrete candidate grasp points \cite{coco} and computing optimal contact forces \cite{boyd_grasp_opt}, which are highly applicable to the planning methods presented here.

Our past work \cite{reachbot_planning,reachbot_sciro} has also explored a similar motion planning problem: given a starting and ending position of the ReachBot body and potential grasp sites, plan a dynamically feasible trajectory and a sequence of stances for minimizing control effort. However, this motion-focused approach does not consider the use of ReachBot for specific manipulation tasks, which typically require a different stance and tensioning than what is produced by the motion planner.


\textit{Statement of Contributions}: This paper presents a manipulation planning method for ReachBot, optimizing for task robustness under uncertain parameters and stochastic failure modes. We establish a two-part architecture: a \textit{stance planner} and a \textit{tension planner}, each inspired by work in the field of dexterous grasping. We present an analysis of how these planners improve performance on relevant tasks, and increase the size of the manipulation workspace. These methods are compared against a baseline method in a Monte-Carlo-based simulation. We also present insight into how these methods perform on similar cable-driven robot architectures. Additional media and code is available at our project webpage, {\small\url{https://stanfordasl.github.io/reachbot_manipulation/}}

\section{Planning Framework}

\begin{figure*}[t]
    \centering
    \smallskip
    \includegraphics[width=\textwidth]{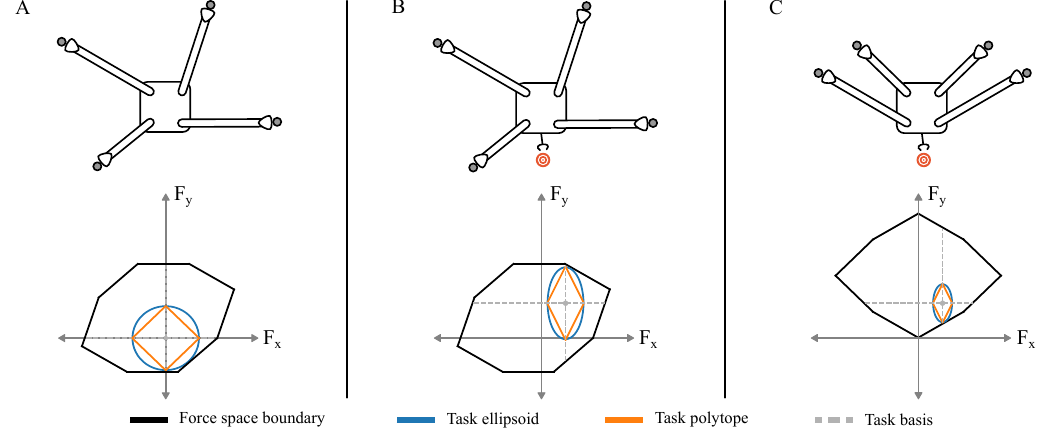}
    \caption{\textbf{Relationship between the ReachBot stance, wrench space, and task polytope.} The X/Y force space, a subset of the wrench space for a 2D ReachBot, is plotted, along with the task polytope and the task ellipsoid. (A): With no specified task, this ReachBot stance can resist arbitrary disturbance forces about the origin of the force space, with magnitudes up to the radius of the ball. (B): With a specified task ellipsoid or polytope, this stance can successfully execute the task even with some uncertainty, since the center of the ellipsoid/polytope is contained in the hull. (C): The task-ellipsoid/polytope metric can also indicate that a task is feasible, even when ReachBot is in a stance which is notably \textit{not} in force closure (i.e. the origin of the space is not contained in the interior of the hull).}
    \label{fig:wrench_space_example}
\end{figure*}

For ReachBot to execute a manipulation task, we consider a multi-stage scenario (Fig. \ref{fig:ops}) where ReachBot will first observe the environment, select the grasp sites to place its booms, and then tension its booms to complete the task. Here, we define a ``task" as a sequence of pose/wrench pairs: a pose in $SE(3)$, and a wrench in $\mathbb{R}^{6}$ (a combination of a force and torque). A single-pose task in the context of ReachBot may imply rejecting disturbances while performing imaging of the environment, and a multi-pose task may involve a pick-and-place task, such as placing sensors or instruments.

To account for these stages, and the different information obtained by the perception system at each stage, we structure the manipulation planner as a two-part system. The \textit{stance planner} uses the perception of the environment (candidate grasp sites and estimated quality of each site) and the task (estimated positioning of the robot and the required wrench/wrenches) to plan a ReachBot stance which can complete the task while being robust against uncertainty and disturbances. Then, the \textit{tension planner} uses the perception of the grasps (their expected pull force distribution based on the microspine gripper model \cite{reachbot_sciro}) to safely tension the booms for the desired task without causing a grasp failure.

\subsection{Assumptions}

To guide the experiments, we assume the following:

\begin{enumerate}
    \item ReachBot is located in an environment where it can span the full width of the space with its booms, such as a lava tube or cave.
    \item We consider an 8-boom configuration, which has been previously shown to satisfy capability constraints while minimizing mass and interference \cite{reachbot_design}.
    \item Manipulation tasks are performed with an end effector on the base of the robot.
    \item All booms are primarily loaded in tension but can support a small bending moment at the shoulder joint. When comparing to CDPRs, we consider cables as pure-tension members.
    \item Booms and cables are massless compared to the body.
\end{enumerate}

We also use the following specifications in our experiments: a ReachBot mass of 10 kg, a maximum gripper pull force of 30N, a maximum boom shoulder moment of 1 Nm, and an environment subject to Martian gravity ($g = 3.71 m/s^2$).

Note that this method can generalize to a parallel robot with any number of cables or booms, even if we have fewer than 7 (the minimum number required for force closure in $\mathbb{R}^{3}$). This method also generalizes to a manipulation task performed at the end of an arm or boom: to do so simply requires a transformation of the resultant wrench back to the base of the robot.


\subsection{Stance planner}

\subsubsection{Preliminaries}


To select the ReachBot stance that maximizes the probability of a successful manipulation task, we leverage tools from dexterous manipulation: namely, wrench spaces, task ellipsoids, and task polytopes \cite{ferrari_canny,sastry_task_ellipsoids,task_quality}. 

Classically, a grasp wrench space is a subset of $\mathbb{R}^{6}$ corresponding to the set of all possible wrenches (forces and torques) that can be applied by a set of point contact forces in a dexterous grasp. We generalize this notion to our setting by considering the set of all possible wrenches that can be applied by the ReachBot body, given its current stance and the maximum force that can be applied along each boom. We use the $L_{\infty}$ definition from \cite{ferrari_canny}, with a modification to describe the unidirectional tensile force from a cable or boom, rather than a cone or pyramid of possible forces from a point contact with friction (Eq. \ref{eq:wrench_space}). Here, $\bigoplus$ denotes the Minkowski sum between the sets of applied wrenches by each boom (each being a line through wrench space), and $w_{i}$ denotes the maximum wrench from boom $i$.
\begin{equation}
    W_{L_{\infty}}=\text {ConvexHull}\left(\bigoplus_{i=1}^n\left\{\mathbf{0}, w_{i}\right\}\right)
    \label{eq:wrench_space}
\end{equation}

Within this wrench space, we can define a task ellipsoid: a set of expected wrenches during the manipulation task, parameterized by an ellipsoid in $\mathbb{R}^{6}$ \cite{sastry_task_ellipsoids}. The sizing of this ellipsoid can be selected to best reflect the task: for instance, in our sample extraction task example (Fig. \ref{fig:rb_manip}), lifting the sample will require forces distributed primarily along the $z$ direction, and we also anticipate some small disturbance forces and torques along other directions. The corresponding ellipsoid, therefore, would be primarily distributed about $z$. 

However, the task ellipsoid is difficult to efficiently optimize for. Instead, we use a task polytope, an $l_1$ approximation to the ellipsoid via its principal axes \cite{task_quality}. We parameterize this task polytope (Eq. \ref{eq:task_polytope}) by a set of task basis wrenches $B_{1}, \ldots, B_{p}$ (each in $\mathbb{R}^{6}$) and weights $\sigma_{1}, \ldots, \sigma_{p}$. The basis $B$ dictates the principal components of the task (typically left as an identity), and the weights $\sigma$ dictate the relative size of the polytope along each basis direction (typically set to the standard deviations of the task in each wrench component).
\begin{equation}
    \mathcal{P}_{task} = \text{ConvexHull}\left(\left\{\sigma_{1} \cdot B_{1}, \ldots, \sigma_{p} \cdot B_{p}\right\}\right)
    \label{eq:task_polytope}
\end{equation}

Using a task polytope also guarantees that all wrenches within its positive spanning set are possible with the given ReachBot stance. Therefore, by selecting an appropriate set of parameters $B$ and $\sigma$, we can guarantee that the task wrench and any uncertain disturbance wrenches are contained in the task polytope and thus can be achieved by the ReachBot stance. This method's independence from force closure enables it to work even when the task is in a difficult location (Fig. \ref{fig:wrench_space_example} C), or if there are a limited number of grasp sites available. This also leads to favorable worst-case outcomes, even in the event of grasp failure and subsequent closure loss.


\subsubsection{Method}

Given a pose and a task ellipsoid defining the distribution of wrenches, we optimize based on the smallest-magnitude disturbance wrench within the task polytope which can be resisted by the ReachBot stance (Eq. \ref{eq:sp_obj}). In general, this objective seeks to maximize the size of the wrench space about the task ellipsoid: an ``inverse Chebyshev ball" problem, where we modify the polyhedron to best fit a ball or ellipsoid. Empirically, optimizing via the task polytope correlates very strongly with the true task ellipsoid.

We also enforce the following constraints:
\begin{itemize}
    \item [(\ref{eq:sp_c1})] \textit{Unique boom assignment}: Each boom can be assigned to at most one grasp site.
    \item [(\ref{eq:sp_c2})] \textit{Unique site assignment}: Each grasp site can be assigned to at most one boom.
    \item [(\ref{eq:sp_c3})] \textit{Wrench space scaling}: The size of the wrench space along each task basis wrench is non-negative.
    \item [(\ref{eq:sp_c4})] \textit{Boom non-interference}: The directions of the booms are each restricted to lie in an outward-facing second-order cone, to prevent self-intersection with the robot body.
    \item [(\ref{eq:sp_c5})] \textit{Tension limits}: Each boom's tension is positive and limited by the maximum gripper force, the quality of the grasp site, and the Boolean assignment variable.
    \item [(\ref{eq:sp_c6})] \textit{Force-wrench space relationship}: This constraint relates the boom forces to the size of the wrench space about the nominal desired wrench. 
\end{itemize}

The stance planner is a mixed-integer convex program (MICP), and can be expressed as follows. Here, $A$ is the boolean boom-to-site assignment matrix, $s$ defines the maximum distance in the wrench space along each basis direction, $D$ is the precomputed set of possible directions between booms and sites, $N$ and $\theta$ parameterize the direction and size of the boom reachability cones, $T$ is the set of boom tensions, $t_{\text{max}}$ is the maximum boom tension, $q$ is an estimated quality parameter for each site, $w$ is a resultant wrench component, $w_{\text{des}}$ is the desired wrench, and $B$ is the set of task basis vectors. $\odot$ denotes the Hadamard (element-wise) product, and ($i$, $j$, $k$) index over all booms, sites, and task basis vectors, respectively.
\begin{align}
        \text {maximize    }     & \min (\sigma^{-1} \odot s)                                                      &                   \label{eq:sp_obj} \\
        \text {subject to    } & A \mathbf{1}_n\leq\mathbf{1}_m                                        &                    \label{eq:sp_c1} \\
                             & A^T \mathbf{1}_m \leq \mathbf{1}_n                                    &                   \label{eq:sp_c2} \\
                             & s \geq \mathbf{0}                                                     &                   \label{eq:sp_c3} \\
                             & D_{ij} \cdot N_i A_{ij} \geq \cos (\theta) A_{ij}                       & \forall i, j   \label{eq:sp_c4} \\
                             & \mathbf{0} \leq T_{ijk} \leq t_{\text{max}} q_j A_{ij}                       & \forall i, j, k   \label{eq:sp_c5} \\
                             & \sum_{i=1}^m \sum_{j=1}^n w_{ijk}=w_{\text{des}}+s_k B_k                     & \forall k \label{eq:sp_c6}
\end{align}

\subsubsection{Handling pose uncertainty}
The task polytope objective inherently optimizes for robustness against wrench uncertainty, yet pose uncertainty is also inherent to this planning problem. For instance, we may estimate a nominal pre-grasp pose for a pick-and-place task, but after approaching the object, it may be better to slightly adjust the pose to best grasp it.

When this occurs, the off-nominal pose will shift the directions of each boom, changing the resultant wrench on the robot's body. Often, these uncertain poses can lead to a dramatic shift in the facets of the nominal wrench space, leading to a potentially much smaller task ellipsoid, or in the worst case, geometric infeasibility of the task. Note that this sensitivity of the wrench space to pose uncertainty is dependent on the configuration of the robot and thus cannot be known prior to evaluating the stance planner.

Unfortunately, expanding the domain of the problem to account for uncertainty in boom directions leads to computational intractability within reasonable planning times. Despite this, we propose a simple method of handling pose uncertainty that integrates easily with our existing wrench-uncertainty-focused planner. 

\begin{figure*}[t]
    \centering
    \smallskip
    \includegraphics[width=\textwidth]{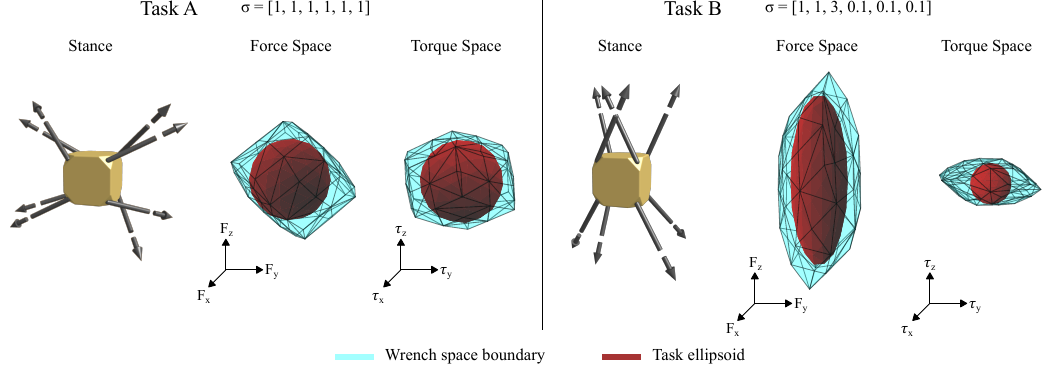}
    \caption{\textbf{Impact of the task-based stance optimization on the ReachBot wrench space.} Here, we show the optimal ReachBot stances and corresponding wrench space (decomposed into the spaces of pure forces and torques) for two different tasks. By changing the task definition from a ball (left) to an ellipsoid biased towards forces along the z direction (right), the \textit{stance planner} adjusts the wrench space to maximize the size of the inscribed task ellipsoid. This enables ReachBot to be robust against uncertainty in the wrench or set of wrenches experienced during manipulation.}
    \label{fig:opt_wrench_spaces}
\end{figure*}

We rely on the empirical observations that (i) orientation errors are the primary contributor to geometric infeasibility of a target wrench, and (ii) stances robust to torque uncertainty are also robust against orientation uncertainty. Using these observations, we can re-scale the torque component of the task polytope, increasing the weighting along these dimensions as a function of orientation uncertainty. Note that when orientation uncertainty is 0, we leave this weighting, $\lambda$ as 1. A simple heuristic that accounts for this is $\lambda = 1 + \lVert{\sigma_{F}}\rVert\lVert{\Delta\Theta}\rVert / \lVert{\sigma_{\tau}}\rVert$, where $\lVert{\Delta\Theta}\rVert$ is the estimated average magnitude of the angular error, and $(\lVert{\sigma_{F}}\rVert, \lVert{\sigma_{\tau}}\rVert)$ are the norms of the force and torque components of the nominal task polytope. As an example, if we expect rotational errors up to 10° in any axis for Task B in Fig. \ref{fig:opt_wrench_spaces}, this scaling factor $\lambda$ equals 4.3, increasing the torque weighting and bringing the task polytope closer to the ball metric shown in Task A.


\subsubsection{Multi-pose extension}
If the manipulation task calls for a sequence of wrenches applied at different poses, we expand the dimensionality of the problem proportionally to the number of poses. For instance, a pick-and-place task may be defined by pose/wrench pairs at four key points: pre-grasp, post-grasp, pre-place, and post-place. Additional pose/wrench pairs can also be added at intermediate positions for long-horizon trajectories. Given this, we then optimize for the smallest-magnitude disturbance wrench across all of these pose/wrench pairs in the trajectory.

\subsection{Tension planner}

Once the stance of ReachBot is fixed, we then determine the tensions in each boom to apply the desired task wrench. For our objective, we employ the probability of success metric from our previous work \cite{reachbot_planning}, based on an estimated force distribution which is generated through a learned model of the grasp quality \cite{reachbot_sciro}.

This convex program (CP) is subject to just two constraints:

\begin{enumerate}
    \item \textit{Tension limits}: Each boom tension is positive and limited by the maximum gripper force and the quality of the grasp site
    \item \textit{Task execution}: The resultant wrench from all booms equals the desired task wrench.
\end{enumerate}

This problem can be expressed as follows, where $\Phi$ represents the Gaussian CDF, $\mu_{i}$, $\sigma_{i}$ represent the estimated pull force normal distribution parameters for the $i^{th}$ grasp, and $t$ is the boom tension variable. As with the stance planner, $t_{\text{max}}$ is the maximum boom tension, $q$ is the site quality parameter, $w_{i}$ is the resultant wrench from the $i^{th}$ boom, and $w_{\text{des}}$ is desired wrench:
\begin{equation*}
    \begin{array}{cl}
        \text {maximize}     & \sum_{i=1}^n \log \left[\Phi\left(\frac{\mu_{i} - t_{i}}{\sigma_{i}}\right)\right] \\
        \text { subject to } & \mathbf{0} \leq t \leq t_{\text{max}} \cdot q                                            \\
                             & \sum_{i=1}^m w_i=w_{\text{des}}
    \end{array}
\end{equation*}

\section{Experiments and Results}

\subsection{Naïve baseline}

\begin{figure}
    \centering
    \includegraphics[width=\linewidth]{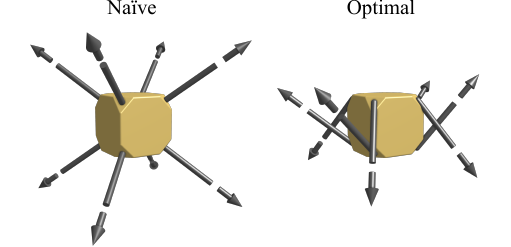}
    \caption{\textbf{Comparison of two possible ReachBot stances for a set of possible sites.} Left: a \textit{naïve} configuration, simply directing the booms outwards from the body. This configuration has good control over the applied force, but is very weak in applying torque, since each boom has no effective lever arm on the body. Right: an \textit{optimal} configuration, with good control over both applied force and torque.}
    \label{fig:naive_opt}
    \vspace{-3mm}
\end{figure}

As a point of comparison for the stance planner, we also develop a naïve baseline method, which extends out the cables (Fig. \ref{fig:naive_opt}) and guarantees that the nominal wrench can be achieved. This method retains many of the same constraints as in the stance planner, but makes no assumptions about robustness to off-nominal wrenches or poses. Notation follows from the stance planner.
\begin{equation*}
    \begin{array}{cll}
        \text {maximize}     & \sum_{i}\sum_{j} A_{ij} D_{ij} \cdot N_{ij}                                                      &                    \\
        \text { subject to } & A \mathbf{1}_n\leq\mathbf{1}_m                                        &                    \\
                             & A^T \mathbf{1}_m \leq \mathbf{1}_n                                    &                    \\
                             & D_{ij} \cdot N_i A_{ij} \geq \cos (\theta) A_{ij}                     & \forall i, j    \\
                             & \mathbf{0} \leq T_{ij} \leq t_{\text{max}} q_j A_{ij}                        & \forall i, j    \\
                             & \sum_{i=1}^m \sum_{j=1}^n w_{ij}=w_{\text{des}}
    \end{array}
\end{equation*}


\subsection{Computation times}

The stance planner, solved with MOSEK via the CVXPY interface, typically finds the optimal solution within 3 seconds for small problems, and around 30 seconds for large problems (see Table \ref{tab:compute}). Here, problem size is dictated by the number of feasible sites for each boom -- a function of the total number of sites, and the cone direction constraint. This compute time is reasonable given ReachBot's speed.

The tension planner, solved with ECOS via the CVXPY interface, typically solves in about 1 millisecond. This is always a fixed problem size, and can be efficiently re-solved online for various applied wrenches.

The naïve planner, also solved with MOSEK, is still a MICP like the stance planner, but its reduced dimensionality allows for significantly faster computation times.

All timings are based on a desktop computer with an Intel Xeon E5-2643 CPU at 3.40GHz, with 64 GB RAM. 
\begin{table}[ht]
    \caption{Average compute times (seconds)}
    \label{tab:compute}
    \centering
    \begin{tabular}{@{}ccccc@{}}
        \toprule
        \multicolumn{1}{l}{}     &          & \multicolumn{2}{c}{Stance planner}  & Tension planner \\ \midrule
        \multicolumn{1}{l}{}     &          & Optimal      & Naïve       &                 \\
        \multirow{3}{*}{Sites}   & 10       & 2.24         & 0.204       & 0.0014          \\
                                 & 20       & 8.10         & 0.244       & 0.0014          \\
                                 & 30       & 29.3         & 0.292       & 0.0014          \\ \bottomrule
    \end{tabular}
    \vspace{-3mm}
\end{table}

\subsection{Evaluation of optimal stance planning on workspace size}

\begin{figure*}[ht]
    \centering
    \smallskip
    \includegraphics[width=\textwidth]{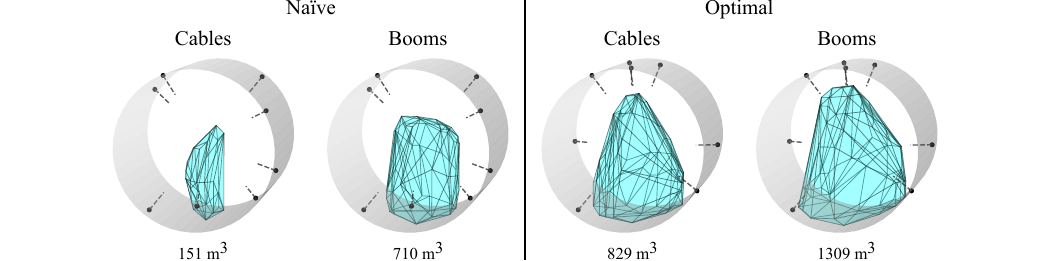}
    \caption{\textbf{Impact of stance optimization on workspace size.} Left: manipulation workspace for cable-driven and boom-driven parallel robots, in a naïve stance. Right: the manipulation workspace for each robot, but in an optimal stance. This stance optimization leads to significant increases in the workspace volume, and reduces the downsides of a lower-degree-of-freedom cable-driven architecture.}
    \label{fig:workspaces}
\end{figure*}

The capability of applying a target wrench is only one part of the manipulation problem: another key requirement is the workspace size. Our previous ReachBot works have already highlighted the large workspace size for ReachBot \cite{reachbot_main}, but this concerns a \textit{reachability} workspace, rather than a \textit{manipulation} workspace. While ReachBot may be able to access a large space with the existing motion planner, there are no guarantees that it can apply the wrenches needed for a manipulation task. In general, we need to see that ReachBot can both access a location (the location is in the \textit{Static Equilibrium Workspace}) and apply the desired wrench (the location is in the \textit{Wrench Closure Workspace}, for that specific wrench) \cite{cdpr_workspaces}.

Therefore, to evaluate the impact of our planner on the workspace, we compare the manipulation workspaces between a \textit{naïve} stance and an \textit{optimal} stance (See Fig. \ref{fig:naive_opt} for a comparison of these stances). Additionally, we look at how the workspace varies between a cable-driven and a boom-driven ReachBot, to help guide the system design as well as demonstrate the applicability of this method to other reconfigurable parallel robots.

As seen in Fig. \ref{fig:workspaces}, in the naïve stance, the cable-driven ReachBot is significantly hindered by the suboptimal cable placement, primarily due to the lack of torque authority in this stance. As purely tensile members, cables must rely solely on the configuration's geometry to achieve a target wrench, whereas booms can passively resist small torques through the reaction torques at their shoulder joints. This can be seen via the torque component of the robot Jacobian: the geometry of a cable-driven robot may lead to this being rank-deficient, while the added degrees of freedom from a boom-driven robot can mitigate this issue. This leads to a relatively large workspace for booms (Fig. \ref{fig:workspaces}), even with a suboptimal stance.

However, if we use the stance planner, this leads to a dramatic improvement in the workspace volume. Both the boom-driven and cable-driven robots see an improvement, though this increase is much more significant for the cable-driven ReachBot. So, while cables are in general less capable than booms, smart planning of the cable positioning mitigates these downsides and leads to comparable performance.

\subsection{Validation in probabilistic simulation}

\begin{figure*}
    \centering
    \smallskip
    \includegraphics[width=\textwidth]{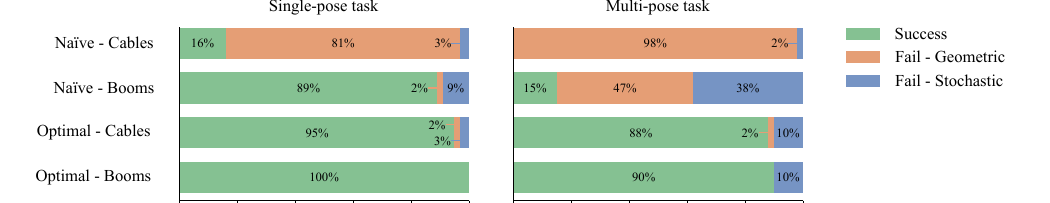}
    \caption{\textbf{Task performance results}, evaluated across randomized environments and tasks in Monte-Carlo simulation. We consider a representative single-pose task as holding a grasped object steady while rejecting arbitrary disturbances, and a multi-pose task as picking-and-placing an object between two locations in the environment. A geometric failure indicates that the task wrench is infeasible given the current stance, and a stochastic failure indicates that a commanded tension exceeds a sampled maximum force from a probabilistic model.}
    \label{fig:results}
\end{figure*}

Our optimization objectives for the stance planner and the tension planner are designed to maximize robustness and safety, but to validate these, we must evaluate the “true” simulated task robustness of these plans. Specifically, we want to show that by using these proposed methods, ReachBot can: (i) apply the required wrench (or set of wrenches) for the manipulation task, even when these can differ from the original plan, (ii) maintain stability even under unmodeled vibrational effects or disturbances, and (iii) execute the task in a manner which leads to a low chance of grasp failure.

To show this, we performed a Monte-Carlo validation of the task robustness in simulation. For 1000 sampled tasks across 10 randomized environments, we determine if (1) the task is geometrically feasible given the planned ReachBot stance, and if so, (2) tension the ReachBot booms and sample from the stochastic grasp site model to determine if a grasp failure occurred. We denote a failure at step (1) as a geometric failure, and a stochastic failure at step (2). Note that stochastic failure is only evaluated if the problem is geometrically feasible (i.e. the desired wrench or set of wrenches lie within the wrench space at each pose).


As seen in Fig. \ref{fig:results}, while the naïve planner is guaranteed to support the nominal task, its lack of uncertainty handling leads to poor performance on off-nominal poses and wrenches. Notably, the main failure mode for the naïve planner is geometric: i.e., given the ReachBot stance, there is no set of boom/cable tensions that will achieve the target wrench. The primary reason for this failure tends to be an inability to produce the required torque for the task, even if this is small. This is especially an issue with cables, which cannot passively resist small torques, unlike booms. This passive torque resistance allows a naïve stance with booms to perform adequately on simple tasks, though as the task complexity increases, performance degrades significantly.

Conversely, our optimal stance planner demonstrates high task performance across both simple and complex tasks, and the stance optimization is particularly critical for these more complex sequences of wrenches and poses. We also see that stochastic failure is generally always present, due to the nature of microspine grasping, but a good configuration can also mitigate the chance of failure. For example, a naïve stance might evenly distribute tensile loads across all sites for a nominal wrench, but rejecting a disturbance may require a very large tensile force applied to a single boom. An optimal stance under the same disturbance will be able to distribute the added loading across multiple booms, reducing the chance of failure.

\section{Conclusion}

In this work, we have established a two-part optimization-based planning method for robust manipulation with ReachBot, leveraging concepts from dexterous grasping and cable-driven parallel robots. First, the stance planner selects the placement of ReachBot's booms through a mixed-integer convex program, adjusting the wrench space to improve a task-polytope-based robustness metric. Second, the tension planner determines the forces in each boom to apply a desired task wrench, solving a fast convex program to minimize the chance of a grasp failure.

Through our experiments, we've shown that these optimizations significantly increase the size of the manipulation workspace, and lead to robust task execution even under disturbances or uncertainty in the desired poses or wrenches required for the task. Given the high reliability of these methods, this new manipulation capability will motivate new missions and scientific experiments on the Moon, Mars, and beyond.

Future work in this area includes integrating these optimal manipulation stances into the ReachBot motion planner, such that the final stance from the motion plan is the optimal stance for the given manipulation task. Second, construction on new reconfigurable cable-driven ReachBot hardware is underway, and experiments on this platform will further demonstrate the need for the planning methods presented here.


\bibliographystyle{ieeetr}
\bibliography{references}

\end{document}